\documentclass[10pt,twocolumn,letterpaper]{article}

\usepackage{cvpr}              

\usepackage{graphicx}
\usepackage{amsmath}
\usepackage{amssymb}
\usepackage{booktabs}

\usepackage{multirow}
\usepackage{pifont}
\graphicspath{{figures/}}
\usepackage{diagbox}
\usepackage{gnuplottex}
\usepackage{float}
\usepackage{bm}

\usepackage[pagebackref,breaklinks,colorlinks]{hyperref}

\usepackage[capitalize]{cleveref}
\crefname{section}{Sec.}{Secs.}
\Crefname{section}{Section}{Sections}
\Crefname{table}{Table}{Tables}
\crefname{table}{Tab.}{Tabs.}

\def\cvprPaperID{9493} 
\def\confName{CVPR}
\def\confYear{2023}

\begin{document}

\title{Robust Mean Teacher for Continual and Gradual Test-Time Adaptation}

\author{
Mario D\"obler \thanks{Equal contribution.} \qquad  Robert A. Marsden \footnotemark[1] \qquad Bin Yang \\
University of Stuttgart\\
{\tt\small  \{mario.doebler, robert.marsden, bin.yang\}@iss.uni-stuttgart.de}
}

\maketitle

\begin{abstract}
Since experiencing domain shifts during test-time is inevitable in practice, test-time adaption (TTA) continues to adapt the model after deployment. Recently, the area of continual and gradual test-time adaptation (TTA) emerged. In contrast to standard TTA, continual TTA considers not only a single domain shift, but a sequence of shifts. Gradual TTA further exploits the property that some shifts evolve gradually over time. Since in both settings long test sequences are present, error accumulation needs to be addressed for methods relying on self-training. In this work, we propose and show that in the setting of TTA, the symmetric cross-entropy is better suited as a consistency loss for mean teachers compared to the commonly used cross-entropy. This is justified by our analysis with respect to the (symmetric) cross-entropy's gradient properties. To pull the test feature space closer to the source domain, where the pre-trained model is well posed, contrastive learning is leveraged. Since applications differ in their requirements, we address several settings, including having source data available and the more challenging source-free setting. We demonstrate the effectiveness of our proposed method ``robust mean teacher`` (RMT) on the continual and gradual corruption benchmarks CIFAR10C, CIFAR100C, and Imagenet-C. We further consider ImageNet-R and propose a new continual DomainNet-126 benchmark. State-of-the-art results are achieved on all benchmarks. \footnote{Code is available at: \url{https://github.com/mariodoebler/test-time-adaptation}}
\end{abstract}

\section{Introduction}
Assuming that training and test data originate from the same distribution, deep neural networks achieve remarkable performance. In the real world, this assumption is often violated for a deployed model, as many environments are non-stationary. Since the occurrence of a data shift \cite{quinonero2008dataset} during test-time will likely result in a performance drop, domain generalization aims to improve robustness and generalization already during training \cite{hendrycks2019augmix, hendrycks2021many, muandet2013domain, tobin2017domain, tremblay2018training}. However, these approaches are often limited, due to the wide range of potential data shifts \cite{mintun2021interaction} that are unknown during training. To gain insight into the current distribution shift, recent approaches leverage the test samples encountered during model deployment to adapt the pre-trained model. This is also known as test-time adaptation (TTA) and can be done either offline or online. While offline TTA assumes to have access to all test data at once, online TTA considers the setting where the predictions are needed immediately and the model is adapted on the fly using only the current test batch. 

While adapting the batch normalization statistics during test-time can already significantly improve the performance \cite{schneider2020improving}, more sophisticated methods update the model weights using self-training based approaches, like entropy minimization \cite{TENT}. However, the effectiveness of most TTA methods is only demonstrated for a single domain shift at a time. Since encountering just one domain shift is very unlikely in real world applications, \cite{CoTTA} introduced \textit{continual test-time adaptation} where the model is adapted to a sequence of domain shifts. As pointed out by \cite{CoTTA}, adapting the model to long test sequences in non-stationary environments is very challenging, as self-training based methods are prone to error accumulation due to miscalibrated predictions. Although it is always possible to reset the model after it has been updated, this prevents exploiting previously acquired knowledge, which is undesirable for the following reason: While some domain shifts occur abruptly in practice, there are also several shifts which evolve gradually over time \cite{kumar2020understanding}. In \cite{GTTA}, this setting is denoted as \textit{gradual test-time adaptation}. \cite{kumar2020understanding, GTTA} further showed that in the setting of gradual shifts, pseudo-labels are more reliable, resulting in a better model adaptation to large domain gaps. However, if the model is reset and the domain gap increases over time, model adaptation through self-training or self-supervised learning may not be successful \cite{kumar2020understanding, liu2021ttt++}. 

To tackle the aforementioned challenges, we introduce a robust mean teacher (RMT) that exploits a symmetric cross-entropy (SCE) loss instead of the commonly used cross-entropy (CE) loss to perform self-training. This is motivated by our findings that the CE loss has undesirable gradient properties in a mean teacher framework which are compensated for when using an SCE loss. Furthermore, RMT uses a multi-viewed contrastive loss to pull test features towards the initial source space and learn invariances with regards to the input space. While our framework performs well for both continual and gradual domain shifts, we observe that mean teachers are especially well suited for easy-to-hard problems. We empirically demonstrate this not only for gradually shifting test sequences, but also for the case where the domain difficulty with respect to the error of the initial source model increases. Since source data might not be available during test-time due to privacy or accessibility reasons, recent approaches in TTA focus on the source-free setting. Lacking labeled source data, source-free approaches can be susceptible to error accumulation. Therefore, as an extension to our framework, we additionally look into the setting where source data is accessible.

We summarize our contributions as follows:
\begin{itemize}
    \item By analyzing the gradient properties, we motivate and propose that in the setting of TTA, the symmetric cross-entropy is better suited for a mean teacher than the commonly used cross-entropy.
    \item We present a framework for both continual and gradual TTA that achieves state-of-the-art results on the existing corruption benchmarks, ImageNet-R, and a new proposed continual DomainNet-126 benchmark.
    \item For our framework, we address a wide range of practical requirements, including the source-free setting and having source data available.
\end{itemize}

\section{Related Work}
\paragraph{Domain Generalization}
Generalizing to unseen test distributions is the area which is generally addressed by domain generalization. It has been shown that data augmentation \cite{shorten2019survey} during training improves generalizability and robustness \cite{hendrycks2019augmix, hendrycks2021many, li2021feature}. Another direction is to learn domain-invariant features, as addressed by \cite{muandet2013domain, dou2019domain}. The idea of domain randomization \cite{tobin2017domain, tremblay2018training} uses different synthesis parameters of simulation environments during the learning process to improve generalization.

\paragraph{Unsupervised Domain Adaptation (UDA)}
Since generalizing to every unseen target domain remains an open question, unsupervised domain adaptation \cite{wilson2020survey} considers the setting, where next to labeled source data, unlabeled target data is also available. This makes the problem easier since the available target data narrows down the domain shift. To increase the performance on the target domain, one line of work focuses on minimizing the discrepancy between the source and target feature distributions, by using either adversarial learning \cite{ganin2016domain, tzeng2017adversarial}, discrepancy based loss functions \cite{chen2020homm, sun2016deep, yan2017mind}, or contrastive learning \cite{kang2019contrastive, marsden2022contrastive}. While it is also possible to adapt the domains in the input space \cite{sankaranarayanan2018generate, CYCADA, ACE, marsden2022continual}, self-training has emerged as a powerful technique. It uses the network's predictions as pseudo-labels to minimize the (cross)-entropy for target data \cite{IAST, ADVENT, liu2021cycle, CRST}. To increase the reliability of the network's predictions, \cite{tranheden2021dacs, french2018selfensembling} rely on a mean teacher \cite{tarvainen2017mean}.

\paragraph{Test-time Adaptation (TTA)}
Test-time adaptation methods adapt a pre-trained model after deployment leveraging the currently available test data. Since the test samples also provide insights into the distribution shift, \cite{schneider2020improving} showed that simply adapting the batch normalization (BN) statistics during test-time can already significantly improve the performance on corrupted data. This is in spirit to \cite{AdaBN} which previously proposed to update the BN statistics in the setting of UDA. While this strategy only requires a forward pass, current approaches in TTA further perform a backward pass in which the model weights are updated. For example, \cite{TENT} minimizes the entropy with respect to the BN parameters. \cite{MEMO} minimizes the entropy with respect to all parameters and uses test-time augmentation \cite{krizhevsky2009learning} to artificially increase the batch size. Other methods apply contrastive learning \cite{chen2022contrastive} or even introduce an additional self-supervision loss during source pre-training that is later leveraged to perform the adaptation during test-time \cite{TTT, liu2021ttt++, MT3, bartler2022ttaps}. Diversity regularizers \cite{liang2020we, mummadi2021test} are a recent approach to circumvent collapse to trivial solutions potentially caused by confidence maximization. While many methods assume to have a batch of test data available, one line of work focuses on single-sample TTA \cite{mirza2022norm, gao2022back, MEMO, MT3}.

\paragraph{Continual and Gradual Test-time Adaptation}
Test-time adaptation methods typically adapt the model to a single target domain. In the real world, encountering a sequence of domain shifts is much more likely. Therefore, continual test-time adaptation considers the case of online TTA with sequentially changing domains. Some of the existing TTA methods can also be applied in the continual setting, such as the online version of TENT \cite{TENT}, but methods solely based on self-training can be prone to error accumulation. This is a result of miscalibrated predictions, as highlighted in \cite{CoTTA}. CoTTA \cite{CoTTA} was the first method specifically proposed for the setting of continual TTA. It uses weight and augmentation-averaged predictions to address the problem of error accumulation, which are further combined with a stochastic restore to circumvent catastrophic forgetting \cite{mccloskey1989catastrophic}. \cite{GTTA} proposes a method that not only considers a continual setting, but also investigates scenarios where shifts are gradual. Based on the theory of \cite{kumar2020understanding} that self-training is sufficient as long as the encountered shifts are small enough, GTTA \cite{GTTA} introduces intermediate domains by mixup or style transfer for effective self-training.

\section{Methodology}
\begin{figure}[t]
\centering
\def\svgwidth{200pt}    
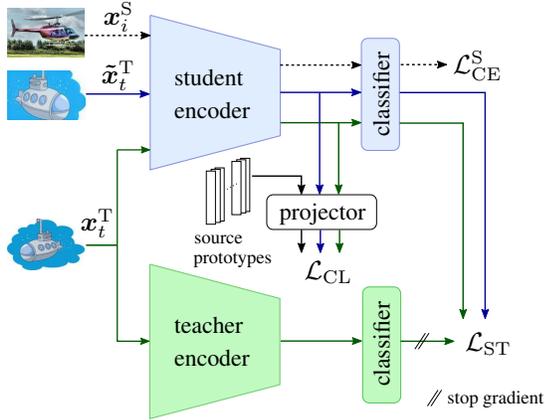  
\caption{\textbf{RMT framework:} Before the adaptation, the student and teacher networks are initialized with source pre-trained weights. Source prototypes are extracted for each class and a mean teacher warm-up is performed. During test-time, the current test batch $\bm{x}_t^\mathrm{T}$ and an augmented version $\Tilde{\bm{x}}_t^\mathrm{T}$ are encoded by the student. Test features, augmented test features, and the nearest source prototypes based on the cosine similarity are used for the contrastive loss $\mathcal{L}_\mathrm{CL}$. Self-training is performed by using two symmetric cross-entropy losses. If source data is available, we uniformly sample a batch $(\bm{x}_i^\mathrm{S}, \bm{y}_i^\mathrm{S})$ from the source data and compute the cross-entropy loss $\mathcal{L}_{\mathrm{CE}}^\mathrm{S}$. The student is updated by minimizing the previously mentioned losses. The teacher is updated via an exponential moving average of the student's parameters.}
\label{fig:framework}
\end{figure}
In many practical applications, the environmental conditions change over time. Hence, a model $f_{\bm\theta_0}$, pre-trained on source data $(\mathcal{X}^\mathrm{S}, \mathcal{Y}^\mathrm{S})$, can quickly become sub-optimal and provide only inadequate predictions for the current test data $\bm{x}_t^\mathrm{T}$ at time step $t$. To prevent the model's performance from deterioration, online test-time adaptation updates the model based on the current test data $\bm{x}_t^\mathrm{T}$. Depending on the application, labeled source data $(\mathcal{X}^\mathrm{S}, \mathcal{Y}^\mathrm{S})$ may be available. 

In this work, we propose a robust mean teacher that leverages the symmetric cross-entropy, which we show to have more desirable gradient properties, contrastive learning to become invariant to small changes in the input space and pull test features towards the initial source space, and optionally source replay depending on whether source data is accessible. We empirically show that performing a mean teacher warm-up is further beneficial. An overview of our framework is depicted in \cref{fig:framework}. 

\subsection{Robust Mean Teacher}
\label{sec:rmt}
Conducting unsupervised domain adaptation or test-time adaptation by using the network's predictions as pseudo-labels to update itself has been shown to be very effective. This technique, known as self-training, can only work if reliable pseudo-labels are provided. One possibility to improve the pseudo-labels are mean teachers (MT) \cite{tarvainen2017mean}. By simply averaging the weights of a student model over time, the resulting teacher model provides a more accurate prediction function than the final function of the student \cite{tarvainen2017mean}.

To realize the MT framework, a student and a teacher model are initialized at time step $t=0$ with source pre-trained weights $\bm{\theta}_0$. During test-time, the student $f_{\bm{\theta}_t}$ is updated ($\bm{\theta}_t \rightarrow\bm{\theta}_{t+1}$) by minimizing the cross-entropy (CE)
\begin{equation}
    \mathcal{L}_{\mathrm{CE}}(\bm{q},\bm{p}) = - \sum_{c=1}^C q_{c}\, \mathrm{log}\, p_{c}
    \label{eq:entropy_loss}
\end{equation}
between the softmax predictions of the student $\bm{p}$ and the teacher $\bm{q}$, with $C$ being the total number of classes. Subsequently, the non-trainable weights of the teacher $\bm{\theta}'_t$ are updated using an exponential moving average $\bm{\theta}'_{t+1} = \alpha \bm{\theta}'_t + (1 - \alpha) \bm{\theta}_{t+1}$, where $\alpha$ denotes the momentum of how much the teacher is updated. Due to the exponential moving average, mean teachers are also known to be more stable in changing environments \cite{laskin2020curl}, which is also a desirable property for continual and gradual test-time adaptation.

\paragraph{Undesirable gradient properties} Looking at the binary case of the cross-entropy, the partial derivative with respect to the student's output probability $p$ is given by
\begin{equation}
\frac{\partial\mathcal{L}_{\mathrm{CE}}}{\partial p}=\frac{p-q}{p-p^2},
\end{equation}
where $p$ and $q$ are the student's and teacher's output probabilities, respectively. First, when both student and teacher have the same confidence $p=q \in [0.5, 1.0)$, the resulting gradient is zero. So even if student and teacher agree with the same confidence that a sample belongs to the same class $p=q>0.5$, there will be no progress. Suppose we have a small shift where our classifier is still able to make correct predictions, albeit with less certainty. If we did not update the decision boundary, another small shift could lead to incorrect predictions, impairing self-training. Second, the gradient is highly imbalanced, especially for the case when either the teacher or student is very confident. The gradient explodes when the student is very confident in contrast to the teacher, resulting in the student to become a lot less confident. On the contrary, when the teacher is very confident, but the student is not, learning is limited due to the relatively small gradient. These effects are illustrated in the form of the loss surface in \cref{fig:loss_surface} and the gradient surface in \cref{fig:grad_surface} in the appendix.

\paragraph{Symmetric cross-entropy has better gradient properties}
\begin{figure*}[t]
  \centering
      \begin{tabular}{cc}
        \includegraphics[width=3.3in]{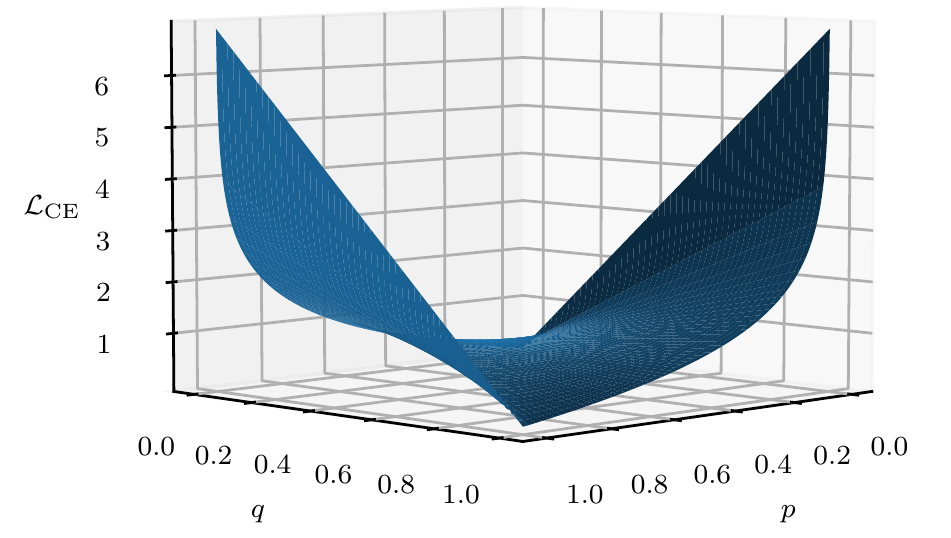} &
        \includegraphics[width=3.3in]{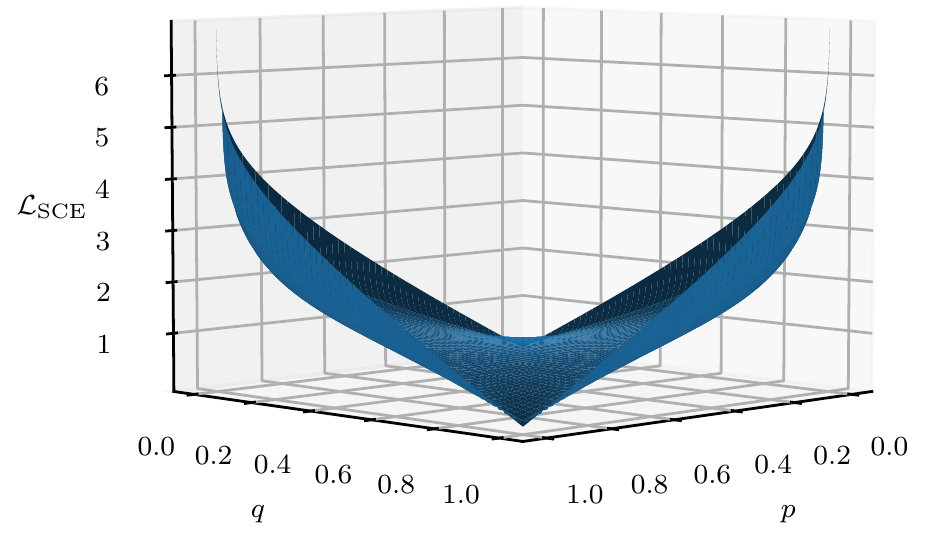}
      \end{tabular}
  \caption{Loss surface illustrated for the binary case of the cross-entropy loss $\mathcal{L}_\mathrm{CE}$ and the symmetric cross-entropy $\mathcal{L}_\mathrm{SCE}$ in dependence of the confidences $p$ and $q$ of the student and teacher, respectively.}
  \label{fig:loss_surface}
\end{figure*}

Next, we propose the usage of the symmetric cross-entropy (SCE) \cite{wang2019symmetric} for a mean teacher and hypothesize that in this setting it has better gradient properties compared to the commonly used cross-entropy loss. The symmetric cross-entropy was originally proposed as a more robust version of the cross-entropy with regards to label-noise \cite{wang2019symmetric}, which is also desirable in our context. For two distributions $\bm{q}$ and $\bm{p}$, the symmetric cross-entropy is defined as 
\begin{equation}
    \mathcal{L}_{\mathrm{SCE}}(\bm{q}, \bm{p}) = - \sum_{c=1}^C q_c\, \mathrm{log}\, p_c - \sum_{c=1}^C p_c\, \mathrm{log}\, q_c,
\end{equation}
where the first term corresponds to the cross-entropy loss $\mathcal{L}_{\mathrm{CE}}$, while the second term is the reverse cross-entropy (RCE) loss $\mathcal{L}_{\mathrm{RCE}}$.

For the binary SCE loss, the gradient is given by
\begin{equation}
    \frac{\partial\mathcal{L}_{\mathrm{SCE}}}{\partial p}=\frac{\partial\mathcal{L}_{\mathrm{CE}}}{\partial p}+\mathrm{log}(1/q-1).
\end{equation}
In the case where the teacher has a probability of $q=0.5$, the derivative of the SCE loss is equivalent to the derivative of the cross-entropy loss. Looking again at the situation where both student and teacher have the same confidence $p=q \in (0.5, 1.0)$, the absolute value of the derivative is larger the more confident the teacher. This leads to increasing the student's confidence, potentially benefiting the self-training process. Moreover, the additional term of the SCE derivative results in a more balanced gradient due to the second term originating from the RCE derivative: $\mathrm{log}(1/q-1)<0$ for $q>0.5$.

Looking at the properties of the cross-entropy loss from another perspective, it is already known that for the CE loss, samples with low confidences dominate the overall gradient \cite{mummadi2021test}. This can be obstructive in the setting of self-training, since low confidence samples are typically more inaccurate. Following the analysis of \cite{chu2022denoised}, the partial derivative of the RCE loss $\mathcal{L}_{\mathrm{RCE}}$ with respect to the $j$-th output element of the network is given by
\begin{equation}
    \frac{\partial \mathcal{L}_{\mathrm{RCE}}}{\partial z_j} = p_j \Big( \sum_{c=1}^C p_c\, \mathrm{log}\, q_c - \mathrm{log}\, q_j \Big).
\end{equation}
Now, if the probability vector $p$ is kept fixed, it becomes apparent that the gradient reaches its maximum when $q$ is a one-hot vector, while the minimum is obtained for a uniform probability vector. Therefore, the reverse cross-entropy loss $\mathcal{L}_{\mathrm{RCE}}$ maintains the performance on high-confidence samples. As a result, we find that the SCE keeps the gradient for high and low confidence predictions balanced, benefiting the optimization problem.

Thus, we now replace the CE loss from \cref{eq:entropy_loss} by the symmetric cross-entropy, which is calculated using the softmax predictions of the teacher $\bm{q}_t^\mathrm{T}$ and student $\bm{p}_t^\mathrm{T}$ for the current test data $\bm{x}_t^\mathrm{T}$. To further promote the consistency between the teacher and student for smaller perturbations, we additionally generate an augmented version of the current test data $\tilde{\bm{x}}_t^\mathrm{T} = \mathrm{Aug}(\bm{x}_t^\mathrm{T})$ using the augmentations from \cite{CoTTA}. $\tilde{\bm{x}}_t^\mathrm{T}$ is then fed through the student network, which provides the softmax prediction $\tilde{\bm{p}}_t^\mathrm{T}$. This prediction is subsequently used to calculate a second SCE loss $\mathcal{L}_{\mathrm{SCE}}(\bm{q}_t^\mathrm{T}, \tilde{\bm{p}}_t^\mathrm{T})$, resulting in the following self-training loss 
\begin{equation}
    \mathcal{L}_{\mathrm{ST}}(\bm{x}_t^\mathrm{T}, \tilde{\bm{x}}_t^\mathrm{T}) = \frac{1}{4}\Big( \mathcal{L}_{\mathrm{SCE}}(\bm{q}_t^\mathrm{T}, \bm{p}_t^\mathrm{T}) + \mathcal{L}_{\mathrm{SCE}}(\bm{q}_t^\mathrm{T}, \tilde{\bm{p}}_t^\mathrm{T}) \Big).
\end{equation}
While it is common to only use the prediction of the teacher as the final output, we ensemble the predictions of both models by adding the student's and teacher's logits. This is motivated by the circumstance that the student can account for distribution shifts more quickly than the slower teacher. Although it is also possible to update the teacher faster, this would affect the teachers ability to stabilize the training.

\paragraph{Mean teacher warm-up for more accurate predictions}
Since in the setting of online adaptation, it takes some time before weight-averaging results in more accurate predictions, we look into performing a mean teacher warm-up before deploying and adapting the model. Warm-up is conducted offline with the same batch size used during test-time by minimizing $\mathcal{L}_\mathrm{SCE}$ for one epoch on $50,000$ source training samples. We want to note that no augmentation is applied during warm-up.

\subsection{Contrastive Learning}
The usage of contrastive learning in our approach is two-fold. On the one hand, it enables to further leverage the augmented test data to learn an invariance to small changes in the input space. On the other hand, the idea is to pull the test feature space towards the source domain where our source pre-trained model is well-posed. 

Before performing test-time adaptation, we first use the pre-trained student encoder at time step $t=0$ to extract a prototype $\bm{r}_{c}^\mathrm{S}$ for each class $c$ in the source dataset. This is achieved by simply averaging all source features belonging to class $c$. Since the prototypes are kept fixed, no source data is required during test-time. 

Now, given test feature $\bm{r}_{ti}^\mathrm{T} = \mathrm{Enc}(\bm{x}_{ti}^\mathrm{T})$ extracted by the student encoder for the $i$-th test image contained in the current test batch with $N$ samples, we first compute the cosine similarity to each source prototype $\bm{r}_{c}^\mathrm{S}$. Then, the nearest source prototype is utilized to create a positive pair, which will later pull each test feature closer to the matching source class center. To further become invariant to small changes in the input space, each pair is extended with the corresponding features of the augmented test batch $\tilde{\bm{x}}_t^\mathrm{T}$. The batch now consists of $3N$ samples due to the test samples, the augmented view, and the nearest source prototypes. Let $i\in I:=\{1,\dots,3N\}$, $A(i):=I\backslash\{i\}$, $V(i)$ be the different views for current sample $i$, and $\bm{z}=Proj(\bm{r})$ denote the output of the non-linear projection of $\bm{r}$, the contrastive loss is then defined as
\begin{equation}
    \mathcal{L}_{\mathrm{CL}} = -\sum_{i \in I} \sum_{v \in V(i)} \mathrm{log} \frac{\mathrm{exp}\big(\mathrm{sim}(\bm{z}_i, \bm{z}_v) / \tau\big)}{\underset{a \in A(i)}{\sum}\mathrm{exp}\big(\mathrm{sim}(\bm{z}_i, \bm{z}_a) / \tau\big)},
    \label{eq:contrastive_loss}
\end{equation}
where $\tau$ denotes the scalar temperature and $\mathrm{sim}(\bm{u},\bm{v})=\bm{u}^T \bm{v}/(\Vert \bm{u}\Vert\Vert \bm{v}\Vert)$ is the cosine similarity.

\subsection{Source Replay}
Inspired by experience replay \cite{lin1992self}, rehearsal is a common technique in continual learning to keep a model in the same low-loss region from which learning was initiated while it is being updated on a new target distribution \cite{verwimp2021rehearsal}. This is also desirable for continual test-time adaptation, since the model may again encounter samples originating from the source distribution or a closely related distribution during test-time. We further want to emphasize that self-training with noisy pseudo-labels can be prone to error accumulation. Even though the mean teacher in combination with the robust symmetric cross-entropy loss already addresses this issue, source replay can also be seen as a stabilizing component for the self-training process, potentially further preventing error accumulation. 

To integrate source replay, we uniformly sample a labeled source batch $(\bm{x}_i^\mathrm{S}, \bm{y}_i^\mathrm{S})$ and minimize
\begin{equation}
    \mathcal{L}_{\mathrm{CE}}^\mathrm{S}(\bm{x}_i^\mathrm{S}, \bm{y}_i^\mathrm{S}) = - \sum_{c=1}^C y_{ic}^\mathrm{S}\, \mathrm{log}\, p_{ic}^\mathrm{S}.
\end{equation}

Clearly, using source replay during test-time requires to store at least parts of the source data in a buffer on the device. Since the buffer size can be a limiting factor, we investigate its influence on the performance later in the experiments. The overall loss function is given as
\begin{equation}
    \mathcal{L}(\bm{x}_t^\mathrm{T}, \tilde{\bm{x}}_t^\mathrm{T}, \bm{x}_i^\mathrm{S}, \bm{y}_i^\mathrm{S}) = \mathcal{L}_\mathrm{ST} + \lambda_\mathrm{CL}\mathcal{L}_\mathrm{CL} + \lambda_\mathrm{CE}\mathcal{L}_\mathrm{CE}^\mathrm{S},
\end{equation}
where the gradient of the loss function is computed with respect to the student's parameters $\bm{\theta}_t$. The teacher $\bm{\theta}'_t$ is updated by an exponential moving average.

\begin{table*}[!h]
\renewcommand{\arraystretch}{1.2}
\centering
\caption{Classification error rate~(\%) for the CIFAR10-to-CIFAR10C, CIFAR100-to-CIFAR100C, and ImageNet-to-ImageNet-C online continual test-time adaptation task on the highest corruption severity level 5. For CIFAR10C the results are evaluated on WideResNet-28, for CIFAR100C on ResNeXt-29, and for Imagenet-C, ResNet-50 is used. We report the performance of our method averaged over 5 runs.} \label{tab:continual-corruptions}
\scalebox{0.93}{
\tabcolsep3pt
\begin{tabular}{l|l|c|c|ccccccccccccccc|c}\hline
& \multicolumn{3}{l|}{Time} & \multicolumn{15}{l|}{$t\xrightarrow{\hspace*{12.2cm}}$}& \\ \hline
& Method & \rotatebox[origin=c]{90}{Source-free} & \rotatebox[origin=c]{90}{Updates} & \rotatebox[origin=c]{70}{Gaussian} & \rotatebox[origin=c]{70}{shot} & \rotatebox[origin=c]{70}{impulse} & \rotatebox[origin=c]{70}{defocus} & \rotatebox[origin=c]{70}{glass} & \rotatebox[origin=c]{70}{motion} & \rotatebox[origin=c]{70}{zoom} & \rotatebox[origin=c]{70}{snow} & \rotatebox[origin=c]{70}{frost} & \rotatebox[origin=c]{70}{fog}  & \rotatebox[origin=c]{70}{brightness} & \rotatebox[origin=c]{70}{contrast} & \rotatebox[origin=c]{70}{elastic} & \rotatebox[origin=c]{70}{pixelate} & \rotatebox[origin=c]{70}{jpeg} & Mean \\
\hline
\multirow{9}{*}{\rotatebox[origin=c]{90}{CIFAR10C}}& Source only & \ding{51} & - & 72.3 & 65.7 & 72.9 & 46.9 & 54.3 & 34.8 & 42.0 & 25.1 & 41.3 & 26.0 & 9.3 & 46.7 & 26.6 & 58.5 & 30.3 & 43.5 \\
& BN--1     & \ding{51} & - & 28.1 & 26.1 & 36.3 & 12.8 & 35.3 & 14.2 & 12.1 & 17.3 & 17.4 & 15.3 & 8.4 & 12.6 & 23.8 & 19.7 & 27.3 & 20.4 \\
& TENT-cont.& \ding{51} & 1 & 24.8 & 20.6 & 28.6 &	14.4 & 31.1 & 16.5 & 14.1 & 19.1 & 18.6 & 18.6 & 12.2 & 20.3 & 25.7 & 20.8 & 24.9 & 20.7\\
& AdaContrast & \ding{51} & 1 & 29.1 & 22.5 & 30.0 & 14.0 & 32.7 & 14.1 & 12.0 & 16.6 & 14.9 & 14.4 & 8.1 & 10.0 & 21.9 & 17.7 & 20.0 & 18.5\\
& CoTTA     & \ding{51} & 1 & 24.3 & 21.3 & 26.6 & 11.6 & 27.6 & 12.2 & 10.3 & 14.8 & 14.1 & 12.4 & 7.5 & 10.6 & 18.3 & 13.4 & 17.3 & 16.2 \\
& GTTA-MIX & \ding{55} & 4 & 23.4 & 18.3 & 25.5 & \textbf{10.1} & 27.3 & 11.6 & 10.1 & 14.1 & 13.0 & 10.9 & 7.4 & 9.0 & 19.4 & 14.5 & 19.8 & 15.6 \\
& RMT (ours)& \ding{51} & 1 & 21.9 & 18.6 & 24.1 & 10.8 & 23.6 & 12.0 & 10.4 & 13.0 & 12.4 & 11.4 & 8.3 & 10.1 & 15.2 & 11.3 & 14.6 & 14.5$\pm$0.09 \\
& RMT (ours)& \ding{55} & 1 & 21.7 & 18.6 & 24.2 & 10.3 & 24.0 & 11.2 & 9.5 & 12.1 & 11.7 & 10.3 & \textbf{7.0} & 8.7 & 14.8 & 10.5 & 14.5 & 13.9$\pm$0.07 \\
& RMT (ours)& \ding{55} & 4 & \textbf{20.8} & \textbf{16.5} & \textbf{20.5} & 10.4 & \textbf{20.1} & \textbf{10.8} & \textbf{9.2} & \textbf{11.0} & \textbf{10.4} & \textbf{9.7} & 7.3 & \textbf{8.0} & \textbf{12.3} & \textbf{8.7} & \textbf{11.6} & \textbf{12.5}$\pm$0.07 \\
\hline
\multirow{9}{*}{\rotatebox[origin=c]{90}{CIFAR100C}}& Source only & \ding{51} & - & 73.0&	68.0&	39.4&	29.3&	54.1&	30.8&	28.8&	39.5&	45.8&	50.3&	29.5&	55.1&	37.2	&74.7&	41.2&	46.4\\
& BN--1     & \ding{51} & - & 42.1 & 40.7 & 42.7 & 27.6 & 41.9 & 29.7 & 27.9 & 34.9 & 35.0 & 41.5 & 26.5 & 30.3 & 35.7 & 32.9 & 41.2 & 35.4 \\
& TENT-cont.& \ding{51} & 1 & 37.2 & 35.8 & 41.7 & 37.9 & 51.2 & 48.3 & 48.5 & 58.4 & 63.7 & 71.1 & 70.4 & 82.3 & 88.0 & 88.5 & 90.4 & 60.9 \\
& AdaContrast & \ding{51} & 1 & 42.3 & 36.8 & 38.6 & 27.7 & 40.1 & 29.1 & 27.5 & 32.9 & 30.7 & 38.2 & 25.9 & 28.3 & 33.9 & 33.3 & 36.2 & 33.4 \\
& CoTTA     & \ding{51} & 1 & 40.1 & 37.7 & 39.7 & 26.9 & 38.0 & 27.9 & 26.4 & 32.8 & 31.8 & 40.3 & 24.7 & 26.9 & 32.5 & 28.3 & 33.5 & 32.5 \\
& GTTA-MIX & \ding{55} & 4 & 36.4 & \textbf{32.1} & 34.0 & \textbf{24.4} & 35.2 & 25.9 & 23.9 & 28.9 & 27.5 & 30.9 & 22.6 & \textbf{23.4} & 29.4 & 25.5
& 33.3 & 28.9 \\
& RMT (ours)& \ding{51} & 1 & 38.5 & 34.4 & 35.4 & 26.4 & 32.7 & 27.0 & 25.0 & 27.5 & 27.6 & 30.0 & 24.0 & 25.8 & 27.0 & 25.2 & 28.4 & 29.0$\pm$0.17 \\
& RMT (ours)& \ding{55} & 1 & 37.4 & 33.8 & 34.3 & 24.8 & 32.0 & 25.3 & \textbf{23.6} & 26.2 & 26.2 & 28.9 & \textbf{21.9} & 23.5 & 25.4 & \textbf{23.2} & 27.4 & 27.6$\pm$0.04 \\
& RMT (ours)& \ding{55} & 4 & \textbf{36.2} & 32.2 & \textbf{32.1} & 25.0 & \textbf{29.8} & \textbf{25.0} & \textbf{23.6} & \textbf{25.4} & \textbf{25.2} & \textbf{27.1} & 23.1 & \textbf{23.4} & \textbf{24.4} & 23.4 & \textbf{25.9} & \textbf{26.8}$\pm$0.08 \\
\hline
\multirow{9}{*}{\rotatebox[origin=c]{90}{ImageNet-C}} & Source only & \ding{51} & - & 97.8 & 97.1 & 98.2 & 81.7 & 89.8 & 85.2 & 78.0 & 83.5 & 77.1 & 75.9 & 41.3 & 94.5 & 82.5 & 79.3 & 68.6 & 82.0 \\
 & BN--1     & \ding{51} & - & 85.0 & 83.7 & 85.0 & 84.7 & 84.3 & 73.7 & 61.2 & 66.0 & 68.2 & 52.1 & 34.9 & 82.7 & 55.9 & 51.3 & 59.8 & 68.6 \\
 & TENT-cont.& \ding{51} & 1 & 81.6 & 74.6 & 72.7 & 77.6 & 73.8 & 65.5 & 55.3 & 61.6 & 63.0 & 51.7 & 38.2 & 72.1 & 50.8 & 47.4 & 53.3 & 62.6 \\
 & AdaContrast & \ding{51} & 1 & 82.9 & 80.9 & 78.4 & 81.4 & 78.7 & 72.9 & 64.0 & 63.5 & 64.5 & 53.5 & 38.4 & 66.7 & 54.6 & 49.4 & 53.0 & 65.5 \\
 & CoTTA     & \ding{51} & 1 & 84.7 & 82.1 & 80.6 & 81.3 & 79.0 & 68.6 & 57.5 & 60.3 & 60.5 & 48.3 & 36.6 & 66.1 & 47.2 & 41.2 & 46.0 & 62.7 \\
 & GTTA-MIX & \ding{55} & 4 & 75.2 & \textbf{67.4} & \textbf{64.6} & 73.3 & 72.5 & 61.8 & \textbf{52.7} & \textbf{53.0} & \textbf{54.9} & \textbf{42.6} & \textbf{33.8} & 63.9 & 48.9 & 44.4 & 47.0 & 57.1 \\
 & RMT (ours)& \ding{51} & 1 & 77.9 & 73.1 & 69.9 & 73.5 & 71.1 & 63.1 & 57.1 & 57.1 & 59.2 & 50.4 & 42.9 & 60.1 & 49.0 & 45.7 & 46.9 & 59.8$\pm$0.18\\
 & RMT (ours)& \ding{55} & 1 & 77.3 & 73.2 & 71.1 & 73.1 & 71.2 & 61.2 & 53.7 & 54.3 & 58.0 & 46.1 & 38.2 & 58.5 & \textbf{45.4} & \textbf{42.3} & \textbf{44.5} & 57.9$\pm$0.26\\
 & RMT (ours)& \ding{55} & 4 & \textbf{74.8} & 68.6 & 65.2 & \textbf{68.2} & \textbf{66.2} & \textbf{59.0} & 53.4 & 53.7 & 56.9 & 47.5 & 41.2 & \textbf{54.1} & 46.0 & 44.6 & 45.9 & \textbf{56.4}$\pm$0.25\\
\hline
\end{tabular}}
\end{table*}
\begin{table*}[!h]
\renewcommand{\arraystretch}{1.1}
\centering
\caption{Classification error rate~(\%) for the gradual CIFAR10-to-CIFAR10C, CIFAR100-to-CIFAR100C, and ImageNet-to-ImageNet-C benchmark averaged over all 15 corruptions. We separately report the performance averaged over all severity levels (@ level 1--5) and averaged only over the highest severity level 5 (@ level 5). The number in brackets denotes the difference to the continual benchmark.}
\label{tab:gradual-corruptions}
\scalebox{0.93}{
\tabcolsep3pt
\begin{tabular}{l|l|cccccccccc}\hline
& & Source & BN--1 & TENT-cont. & AdaCont. & CoTTA & GTTA-MIX & RMT & RMT  & RMT \\ \hline
& Source-free & \ding{51} & \ding{51}  & \ding{51} & \ding{51} & \ding{51} & \ding{55} & \ding{51} & \ding{55} & \ding{55} \\ 
& Updates & - & - & 1 & 1 & 1 & 4 & 1 & 1  & 4 \\ \hline
\multirow{2}{*}{CIFAR10C} & @level 1--5  & 24.7 & 13.7 & 20.4 & 12.1 & 10.9 & 11.8 & 9.3 & \textbf{8.1} & 8.6\\
& @level 5 & 43.5 & 20.4 & 25.1 \small{(+4.4)} & 15.8 \small{(-2.7)} & 14.2 \small{(-2.0)} & 13.0 \small{(-2.6)} & 10.4 \small{(-4.1)} & 9.4 \small{(-4.5)} & \textbf{9.0} \small{(-3.5)} \\ 
\hline
\multirow{2}{*}{CIFAR100C} & @level 1--5 & 33.6 & 29.9 & 74.8 & 33.0 & 26.3 & 24.7  & 26.4 & \textbf{23.6} & 24.2 \\
& @level 5 & 46.4 & 35.4 & 75.9 \small{(+15.0)} & 35.9 \small{(+2.5)} & 28.3 \small{(-4.2)} & 26.1 \small{(-2.8)} & 26.9 \small{(-2.1)} & \textbf{24.3} \small{(-3.2)} & 24.5 \small{(-2.3)} \\ \hline
\multirow{2}{*}{Imagenet-C} & @level 1--5  & 58.4 & 48.3 & 46.4 & 66.3 & 38.8 & 37.7 & 39.3 & 37.8 & \textbf{36.8} \\
& @level 5 & 82.0 & 68.6 & 58.9 \small{(-3.7)} & 72.6 \small{(+7.1)} & 43.1 \small{(-19.6)} & 47.7 \small{(-9.4)} & 41.5 \small{(-18.3)} & 40.2 \small{(-17.7)} & \textbf{37.5} \small{(-18.9)} \\ \hline
\end{tabular}}
\end{table*}

\section{Experiments}

\begin{table}[t] 
\renewcommand{\arraystretch}{1.2}
\centering
\caption{Classification error rate~(\%) for ImageNet-R and DomainNet-126 in the online continual TTA setting. We report the performance of our method averaged over 5 runs.} 
\label{tab:imagenetr_and_domainnet}
\scalebox{0.92}{
\tabcolsep3pt
\begin{tabular}{l|c|c|c||cccc|c}
\hline
\multicolumn{4}{c||}{} & \multicolumn{5}{c}{DomainNet-126} \\
\hline
Method  & \rotatebox[origin=c]{90}{ Source-free } & \rotatebox[origin=c]{90}{ Updates } & \rotatebox[origin=c]{70}{ImageNet-R} &\rotatebox[origin=c]{70}{real $\rightarrow$} & \rotatebox[origin=c]{70}{clipart $\rightarrow$} & \rotatebox[origin=c]{70}{painting $\rightarrow$} & \rotatebox[origin=c]{70}{sketch $\rightarrow$} & Mean \\
\hline
 Source only  & \ding{51} & - & 63.8 & 45.3 & 49.3 & 41.7 & 44.8 & 45.3 \\
 BN--1        & \ding{51} & - & 60.4 & 45.1 & 45.2 & 39.5 & 37.8 & 41.9 \\
 TENT cont.   & \ding{51} & 1 & 57.6 & 42.4 & 44.2 & 37.2 & 37.5 & 40.3 \\
 CoTTA        & \ding{51} & 1 & 57.4 & 43.4 & 43.0 & 36.4 & 36.3 & 39.8 \\
 AdaContrast  & \ding{51} & 1 & 59.1 & 37.8 & 37.6 & 32.3 & 31.9 & 34.9 \\
 GTTA-MIX     & \ding{55} & 4 & 56.6 & 38.7 & 42.4 & 33.6 & 34.2 & 37.2 \\
 RMT (ours)   & \ding{51} & 1 & 55.7 & 37.0 & 37.9 & 31.7 & 32.1 & 34.7 \\
 RMT (ours)   & \ding{55} & 1 & 55.5 & 36.8 & 37.1 & 30.6 & 31.1 & 33.9 \\
 RMT (ours)   & \ding{55} & 4 & \textbf{53.5} & \textbf{35.1} & \textbf{36.4} & \textbf{29.9} & \textbf{29.9} & \textbf{32.8}  \\
\hline
\end{tabular}}
\end{table}

\paragraph{Datasets, metrics, and considered settings}
We evaluate our approach on CIFAR10C, CIFAR100C, and Imagenet-C, which were initially designed to benchmark robustness of classification networks \cite{hendrycks2019benchmarking}. All datasets include 15 different types of corruptions with 5 severity levels applied to the validation and test images of ImageNet and CIFAR, respectively \cite{krizhevsky2009learning}. To validate the effectiveness of our approach for domain shifts not caused by corruption, we additionally consider ImageNet-R \cite{hendrycks2021many}, as well as DomainNet-126 \cite{saito2019semi}, which is a subset of DomainNet \cite{peng2019moment}. While ImageNet-R contains 30,000 examples depicting different renditions of 200 ImageNet classes, DomainNet-126 has 126 classes and consists of four domains (real, clipart, painting, sketch).

We compare all methods in two different settings. First, we consider the continual benchmark, as introduced by \cite{CoTTA}. Similar to the standard TTA setting used in \cite{TENT}, the continual benchmark also starts with an off-the-shelf model pre-trained on the source domain. However, while the standard TTA setting resets the model after it was adapted to a test domain, the continual setting does not assume to know when the domain changes. Therefore, the model is adapted to a sequence of test domains in an online fashion. In case of the corruption benchmark, the sequence consists of all 15~corruptions, each encountered at the highest severity level~5. For DomainNet-126, we randomly create four different domain sequences, with the only condition that every domain is used once as the source domain. More information about the DomainNet-126 benchmark and its benefits are located in \cref{sec:appendix-domainnet}.

Since there are many applications where the domain shift does not occur abruptly but changes rather smoothly, we additionally consider the same gradual benchmark as in \cite{GTTA}. While the continual setting encounters each corruption at the highest severity level 5, the gradual setting increases the severity as follows: $1 \rightarrow 2 \rightarrow \dots \rightarrow 5 \rightarrow \dots \rightarrow 2 \rightarrow 1$. We report the error rate for all experiments.

\paragraph{Implementation details}
Following the RobustBench benchmark \cite{croce2020robustbench}, a pre-trained WideResNet-28 \cite{zagoruyko2016wide} and ResNeXt-29 \cite{xie2017aggregated} is used for CIFAR10-to-CIFAR10C and CIFAR100-to-CIFAR100C, respectively. For ImageNet-to-Imagenet-C, ImageNet-R, and DomainNet-126, a source pre-trained  ResNet-50 is applied. In the latter case, we use the same architecture and pre-trained weights as in \cite{chen2022contrastive}. We follow the implementation of \cite{CoTTA}, using the same hyperparameters. We weight all loss functions equally using $\lambda_\mathrm{CL}=\lambda_\mathrm{CE}=1$ and set $\tau$ to the default value $0.1$.

\paragraph{RMT variations}
Since each application has its own requirements in terms of efficiency, privacy, and memory, we introduce three variations of our method RMT. While the first variant omits source replay to account for situations where it is critical to store source data on the device, the latter two apply source replay but differ in the number of updates. Hence, they address the potential trade-off between efficiency and performance and are meant for applications where memory and computational power are not an issue.

\paragraph{Baselines}
To compare our method, we consider several source-free baselines, such as CoTTA \cite{CoTTA}, TENT continual \cite{TENT}, and AdaContrast \cite{chen2022contrastive}. In addition, we also compare to the non-source-free baseline GTTA-MIX \cite{GTTA} and the normalization-based method BN--1, which recalculates the batch normalization statistics using the current test batch.

\begin{table}[!t] 
\renewcommand{\arraystretch}{1.2}
\centering
\caption{Classification error rate~(\%) for different configurations averaged over 3 runs.} \label{tab:ablation-components}
\scalebox{0.92}{
\tabcolsep3pt
\begin{tabular}{l| ccccc|c}\hline
Method  & \rotatebox[origin=c]{70}{CIFAR10C} & \rotatebox[origin=c]{70}{ CIFAR100C } & \rotatebox[origin=c]{70}{ImageNet-C} & \rotatebox[origin=c]{70}{ImageNet-R} & \rotatebox[origin=c]{70}{DomNet-126} & Mean \\
\hline
MT ($\mathcal{L}_\mathrm{CE}$)           & 18.8 & 32.1 & 65.9 & 59.6 & 41.9 & 43.7 \\
\hline
MT ($\mathcal{L}_\mathrm{SCE}$)         & 17.9 & 31.5 & 62.8 & 57.3 & 39.7 & 41.8 \\
 + warm-up    & 16.7 & 30.6 & 61.3 & 55.0 & 38.8 & 40.5 \\
\hline
 \textbf{A} $\mathcal{L}_{\mathrm{ST}}$ & 18.0 & 31.2 & 61.9 & 57.2 & 39.1 & 41.5\\
 \textbf{B} + ensemble                & 17.1 & 30.5 & 60.1 & 56.5 & 36.8 & 40.2 \\
 \textbf{C} + contrastive             & 16.7 & 30.1 & 59.9 & 55.6 & 35.0 & 39.5 \\
 \textbf{D} + warm-up                 & 14.5 & 29.0 & 59.8 & 55.7 & 34.7 & 38.7 \\
 \textbf{E} + src. replay             & \textbf{13.9} & \textbf{27.6} & \textbf{57.9} & \textbf{55.5} & \textbf{33.9} & \textbf{37.8} \\
\hline
\end{tabular}}
\end{table}

\subsection{Results for Continual Test-Time Adaptation}
\paragraph{Domain shifts caused by corruption} \Cref{tab:continual-corruptions} shows the results for each corruption dataset in the continual setting. While the simple evaluation of the pre-trained source model (source only) leads to a high average error on all datasets, applying test-time normalization with BN--1 already drastically decreases the error rate without any error accumulation. This does not hold for TENT-continual, which suffers from heavy error accumulation for CIFAR100C, achieving an average error of 60.9\%. Nevertheless, it performs on par and 6\% better than BN--1 for CIFAR10C and Imagenet-C, respectively. Although it is always possible to use TENT-episodic, resetting the model after each update prevents the exploitation of previously learned knowledge, resulting in an equivalent performance to BN--1. CoTTA, on the other hand, is able to reduce the average error on most of the datasets without any signs of error accumulation. However, these results are achieved by applying heavy test-time augmentation, requiring up to 32 additional forward passes. If we now compare our source-free variant with CoTTA, the average error is significantly reduced. This variant even outperforms the non-source-free approach GTTA-MIX on CIFAR10C, while being only slightly worse on CIFAR100C. If access to source data is not an issue, the error rate can be further decreased. For applications, where the focus is less on efficiency and more on performance, the error rate can be further reduced by applying 4 update steps, as was also done by GTTA-MIX. Note that applying more update steps to source-free methods usually increases the error rate.

\paragraph{Natural domain shifts} \Cref{tab:imagenetr_and_domainnet} shows the results for ImageNet-R and each sequence included in the continual DomainNet-126 benchmark. As depicted, all methods improve upon the non-adaptive source baseline. While CoTTA performs only slightly better than TENT continual in both settings, AdaContrast clearly takes the lead on DomainNet-126, while lacking performance on ImageNet-R. In contrast, our source-free approach sets new state-of-the-art results on both datasets and is even better than the non-source-free approach GTTA-MIX. If we further leverage source replay, the error rate decreases again, reaching the best results when 4 update steps are applied.

\subsection{Results for Gradual Test-Time Adaptation}
\paragraph{Mean teachers are strong easy-to-hard learners} In \cref{tab:gradual-corruptions}, we report  the average error in the gradual setting across all severity levels and only with respect to level 5. This allows a direct comparison with the continual setting. While the performance of approaches like TENT continual and AdaContrast even degrades for some datasets, mean-teacher based approaches show a massive improvement of more than 18.3\%. Hence, they can exploit the gradual shifts more effectively to reduce the error at level 5. Since a gradual increase in the severity level can also be seen as an easy-to-hard problem, we now revisit the continual setting, but sort the corruptions from low error to high error using the initial source model. Detailed results and the specific sequences are shown in \cref{tab:difficulty} and \ref{tab:seq_descrition_low2high} in the appendix. Again, we find that mean teachers are particularly well suited for easy-to-hard problems, where the error is 12\% lower on ImageNet-C compared to an hard-to-easy sequence.

\subsection{Single-Sample Test-Time Adaptation}
\begin{table}[t]
\renewcommand{\arraystretch}{1.2}
\centering
\caption{Classification error rate~(\%) for single-sample TTA.} \label{tab:sliding_window}
\scalebox{0.93}{
\tabcolsep3pt
\begin{tabular}{l|c|ccccc|c}\hline
Method & \rotatebox[origin=c]{90}{ Window size } & \rotatebox[origin=c]{70}{CIFAR10C} & \rotatebox[origin=c]{70}{CIFAR100C} & \rotatebox[origin=c]{70}{ImageNet-C} & \rotatebox[origin=c]{70}{ImageNet-R} & \rotatebox[origin=c]{70}{DomNet-126} & Mean  \\
\hline
Source only & -  & 43.5 & 46.4 & 82.0 & 63.8 & 45.3 & 56.2 \\
\hline
BN--1       & 8  & 26.3 & 43.8 & 74.6 & 64.7 & 49.7 & 51.8 \\
BN--1       & 16 & 23.2 & 39.5 & 71.0 & 62.2 & 45.4 & 48.3 \\
BN--1       & 32 & 21.9 & 37.4 & 69.3 & 60.7 & 43.4 & 46.5 \\
\hline
RMT (ours)  & 8  & 16.7 & 33.6 & 72.0 & 59.7 & 43.7 & 45.1 \\
RMT (ours)  & 16 & 15.2 & 30.8 & 63.9 & 57.8 & 36.8 & 40.9 \\
RMT (ours)  & 32 & 14.3 & 28.1 & 59.9 & 56.1 & 35.3 & 38.7 \\
\hline
\end{tabular}}
\end{table}

Since timeliness can be important for some applications, we now consider single-sample TTA. A simple approach to overcome noisy gradients and poor estimates of the BN statistics caused by only having a single sample is to use a sliding window. In this case, the last $b$ test samples are stored in a buffer. We only update the model weights every $b$ steps, due to the correlation induced by the buffer. In the meantime, the entire buffer is forwarded to generate a prediction for the current test sample $\bm{x}_{ti}^\mathrm{T}$. Due to the smaller batch size, we decrease the learning rate by $\mathrm{original\ batch\ size}/b$. \Cref{tab:sliding_window} illustrates the results for single-sample TTA using various buffer sizes $b$. Due to the much smaller batch size used in this setting, the performance of the baseline BN--1 slightly degrades as the estimation of the batch statistics becomes more noisy. Although the performance of our approach is also slightly worse compared to the results obtained in the batch setting of TTA, the performance at a window size of 16 is still better or competitive to the state-of-the-art methods in the batch setting.

\subsection{Ablation Studies}
\paragraph{Component analysis}
First, we examine the effect of exploiting the symmetric cross-entropy loss $\mathcal{L}_{\mathrm{SCE}}$. As shown in \cref{tab:ablation-components}, using a mean teacher with $\mathcal{L}_{\mathrm{SCE}}$ has a clear advantage over $\mathcal{L}_{\mathrm{CE}}$. If we further shortly warm up the mean teacher on the source domain using a linear learning rate increase, another significant reduction in error can be achieved on all datasets. Next, we carefully investigate our components. Utilizing our self-training loss $\mathcal{L}_{\mathrm{ST}}$ (A) in combination with the ensemble prediction (B) significantly improves the results compared to the mean teacher framework minimizing either the cross-entropy or the symmetric cross-entropy. While extending our approach with a contrastive component (C) further reduces the average error for all datasets, adding warm-up (D) and source replay (E) again substantially improves the overall performance.

\paragraph{Ablations}
Additional investigations concerning the effect of different numbers of update steps, various amounts of saved source samples, and a sensitivity analysis with respect to the temperature $\tau$ and the momentum term $\alpha$ are shown in \cref{tab:ablations_combined} in the appendix. While we find that 2 and 4 update steps provide a good balance between performance and computational complexity, RMT profits from even more update steps. Although even our source-free variant already sets new standards on all benchmarks, having access to only 1\% of the source data during test-time is beneficial.

\section{Conclusion}
In this work, we showed that a mean teacher with a symmetric cross-entropy loss combined with contrastive learning sets a new standard in the area of continual and gradual TTA. We motivate the usage of a symmetric cross-entropy loss by analyzing the respective gradient properties. We achieve state-of-the-art results on all common benchmarks and introduced a new benchmark based on DomainNet-126 to further demonstrate the effectiveness for a larger variety of domain shifts. In case privacy or accessibility is no concern, replaying a small percentage of source data improves the performance and allows to perform multiple update steps, resulting in an additional performance gain.

\paragraph{Acknowledgments}
This publication was created as part of the research project "KI Delta Learning" (project number: 19A19013R) funded by BMWi.


{\small
\bibliographystyle{ieee_fullname}
\bibliography{egbib}
}


\appendix
\begin{table}[H]
\renewcommand{\arraystretch}{1.2}
\centering
\caption{Classification error rate~(\%) of various methods when the domain difficulty with respect to the initial source error at severity level 5 either increases from low-to-high (\textit{easy-to-hard}) or decreases from high-to-low (\textit{hard-to-easy}).} \label{tab:difficulty}
\scalebox{0.93}{
\tabcolsep4pt
\begin{tabular}{l|c|c|ccc}\hline
Method & \rotatebox[origin=c]{90}{Source-free} & Source error & \rotatebox[origin=c]{70}{CIFAR10C} & \rotatebox[origin=c]{70}{ CIFAR100C } & \rotatebox[origin=c]{70}{ImageNet-C}  \\
\hline
BN--1 & \ding{51} & - & 20.4 & 35.4 & 68.6 \\
\hline
TENT & \ding{51} & high $\rightarrow$ low & 19.6 & 66.9 & 62.8 \\
TENT & \ding{51} & low $\rightarrow$ high & 20.2 & 52.1 & 60.2 \\
\hline
AdaContrast & \ding{51} & high $\rightarrow$ low & 18.8 & 33.7 & 66.3 \\
AdaContrast & \ding{51} & low $\rightarrow$ high & 17.9 & 32.6 & 60.2 \\
\hline
GTTA-MIX & \ding{55} & high $\rightarrow$ low & 17.6 & 30.5 & 60.2 \\
GTTA-MIX & \ding{55} & low $\rightarrow$ high & 17.4 & 30.1 & 58.8 \\
\hline
MT + $\mathcal{L}_{\mathrm{CE}}$ & \ding{51} & high $\rightarrow$ low & 19.2 & 32.4 & 66.4 \\
MT + $\mathcal{L}_{\mathrm{CE}}$ & \ding{51} & low $\rightarrow$ high & 17.1 & 30.5 & 61.2 \\
\hline
MT + $\mathcal{L}_{\mathrm{SCE}}$ & \ding{51} & high $\rightarrow$ low & 18.9 & 31.9 & 63.5 \\
MT + $\mathcal{L}_{\mathrm{SCE}}$ & \ding{51} & low $\rightarrow$ high & 16.7 & 29.4 & 51.5 \\
\hline
RMT (ours) & \ding{55} & high $\rightarrow$ low & 14.2 & 27.7 & 57.5 \\
RMT (ours) & \ding{55} & low $\rightarrow$ high & 12.9 & 26.3 & 50.3 \\
\hline
\end{tabular}}
\end{table}

\begin{table*}[h]
\renewcommand{\arraystretch}{1.2}
\centering
\caption{The corruption types are ordered with respect to the error at severity level 5 of the initial source model from low-to-high.} \label{tab:seq_descrition_low2high}
\scalebox{0.93}{
\tabcolsep3pt
\begin{tabular}{l|lllllllllllllll}\hline
 & \multicolumn{15}{l}{$\text{low}\xrightarrow{\hspace*{14.7cm}}\text{high}$} \\
\hline
CIFAR10C & \rotatebox[origin=c]{51}{ brightness } & \rotatebox[origin=c]{51}{snow} & \rotatebox[origin=c]{51}{fog} & \rotatebox[origin=c]{51}{elastic} & \rotatebox[origin=c]{51}{jpeg} & \rotatebox[origin=c]{51}{motion} & \rotatebox[origin=c]{51}{frost} & \rotatebox[origin=c]{51}{zoom} & \rotatebox[origin=c]{51}{contrast} & \rotatebox[origin=c]{51}{defocus}  & \rotatebox[origin=c]{51}{glass} & \rotatebox[origin=c]{51}{pixelate} & \rotatebox[origin=c]{51}{shot} & \rotatebox[origin=c]{51}{Gaussian} & \rotatebox[origin=c]{51}{impulse} \\
\hline
CIFAR100C & \rotatebox[origin=c]{51}{zoom} & \rotatebox[origin=c]{51}{defocus} & \rotatebox[origin=c]{51}{ brightness } & \rotatebox[origin=c]{51}{motion} & \rotatebox[origin=c]{51}{elastic} & \rotatebox[origin=c]{51}{impulse} & \rotatebox[origin=c]{51}{snow} & \rotatebox[origin=c]{51}{jpeg} & \rotatebox[origin=c]{51}{frost} & \rotatebox[origin=c]{51}{fog} & \rotatebox[origin=c]{51}{glass}  & \rotatebox[origin=c]{51}{contrast} & \rotatebox[origin=c]{51}{shot} & \rotatebox[origin=c]{51}{Gaussian} & \rotatebox[origin=c]{51}{pixelate} \\
\hline
ImageNet-C & \rotatebox[origin=c]{51}{ brightness } & \rotatebox[origin=c]{51}{jpeg} & \rotatebox[origin=c]{51}{fog} & \rotatebox[origin=c]{51}{frost} & \rotatebox[origin=c]{51}{zoom} & \rotatebox[origin=c]{51}{pixelate} & \rotatebox[origin=c]{51}{defocus} & \rotatebox[origin=c]{51}{elastic} & \rotatebox[origin=c]{51}{snow} & \rotatebox[origin=c]{51}{motion}  & \rotatebox[origin=c]{51}{glass} & \rotatebox[origin=c]{51}{contrast} & \rotatebox[origin=c]{51}{shot} & \rotatebox[origin=c]{51}{Gaussian} & \rotatebox[origin=c]{51}{impulse} \\
\hline
\end{tabular}}
\end{table*}

\section{Adaptation with increasing difficulty}
Since mean teacher based approaches have shown tremendous performance improvements in the gradual setting compared to the continual setting, we now consider the case, where the domain difficulty changes from easy-to-hard and hard-to-easy. Specifically, we sort the order of the corruptions with respect to the error at severity level 5 of the initial source model from low-to-high and high-to-low. While \cref{tab:difficulty} shows the results, \cref{tab:seq_descrition_low2high} illustrates the specific corruption orders for an increasing source error. Clearly, all methods improve when the domain difficulty increases compared to the other way round. Notably, mean teachers using a symmetric cross-entropy loss (MT + $\mathcal{L}_\mathrm{SCE}$) demonstrate the highest error reductions with up to 12\% on the ImageNet-C sequence. In contrast, a mean teacher with a cross-entropy loss (MT + $\mathcal{L}_\mathrm{CE}$) only decreases the error by 5.2\%, achieving an error rate of 61.2\%. This is absolutely 9.7\% worse compared to the error rate of a mean teacher using a symmetric cross-entropy loss.

\section{Ablation studies}
\label{sec:appendix_ablatio_studies}
For the following ablation studies, we investigate our non-source-free variant with 1 update step.

\newpage
\paragraph{More updates decrease the error for non-source-free methods}
Since for some applications computational efficiency may be more important than a high accuracy and vice versa, we now investigate the effect of different numbers of update steps. As shown by \cref{tab:ablations_combined} (a), all datasets profit when more update steps are applied, with 2 and 4 steps providing a good balance between performance and computational complexity. However, the best results are achieved with 6 updates. Note that the performance of source-free approaches like CoTTA deteriorates for multiple update steps due to over-adaptation.

\paragraph{1\% of the source data improves the performance}
In \cref{tab:ablations_combined} (b), we illustrate the error rate for different amounts of randomly sampled source data. This is especially relevant for applications with either a limited memory or when source data cannot be stored on the device due to privacy issues (0\%). While the error increases slightly on CIFAR100C and DomainNet-126, the other datasets are only marginally affected by the amount of available source data. Although even our source-free variant already achieves state-of-the-art performance on all benchmarks, storing only 1\% of the source data is enough to further boost the performance.

\paragraph{Sensitivity}
To investigate the sensitivity of our proposed method with respect to the momentum value $\alpha$ of the mean teacher, as well as the temperature term $\tau$, we conduct two ablation studies. As illustrated in \cref{tab:ablations_combined} (c), which shows different values for the temperature term, RMT performs not only stable for all the common default values in contrastive learning (0.07, 0.1 and 0.2), but also for much larger values like 1.0. As shown by \cref{tab:ablations_combined} (c), updating the mean teacher too slow or too fast can slightly degrade the results on average. Nevertheless, for the most common range of momentum values, the performance is stable.

In \cref{tab:appendix_ablation_lambdas}, we analyze the influence of source replay and the contrastive loss by considering different values for $\lambda_{\mathrm{CE}}$ and $\lambda_\mathrm{CL}$, respectively. For values close to the ones used by our approach $\lambda\in[0.5, 1.0]$, we observe a stable performance. As expected, for small values $\lambda\leq0.1$, we observe a drop in performance, underlining that source replay and contrastive learning is indeed beneficial.
\begin{table}[H]
\renewcommand{\arraystretch}{1.2}
\centering
\caption{Classification error rate~(\%) when using different loss weights. The results are averaged over 3 runs and all datasets.} \label{tab:appendix_ablation_lambdas}
\scalebox{0.93}{
\tabcolsep4pt
\begin{tabular}{l|lllll}\hline
$\lambda$ & 0.0 & 0.1 & 0.5 & 1.0 \\
\hline
$\lambda_{\mathrm{CE}}$ (source replay) & 38.7 & 38.1 & 37.6 & 37.8  \\
$\lambda_{\mathrm{CL}}$ (contrastive learning)   & 38.6 & 38.4 & 38.0 & 37.8  \\
\hline
\end{tabular}}
\vskip -0.05in
\end{table}

\begin{table*}[t]
    \renewcommand{\arraystretch}{1.2}
    \centering
    \setlength{\tabcolsep}{20pt}
    \caption{Classification error rate~(\%) for different: (a) numbers of update steps; (b) amounts of available source samples during test-time; (c) temperatures $\tau$ for the contrastive loss; (d) momentum values $\alpha$ used to update the mean teacher.}
    \begin{tabular}{cc}
        (a) & (b) \\
        \scalebox{0.93}{
            \tabcolsep4pt
            \begin{tabular}{l|ccccc}\hline
            Updates & 1 & 2 & 4 & 6 & 8 \\
            \hline
            CIFAR10C       & 13.9 & 13.2 &         12.5  &         12.0  & \textbf{11.8} \\
            CIFAR100C      & 27.6 & 26.9 & \textbf{26.8} & \textbf{26.8} &         27.0  \\
            ImageNet-C     & 57.9 & 56.9 &         56.4  & \textbf{56.1} &         56.4  \\
            ImageNet-R     & 55.5 & 54.6 &         53.5  & \textbf{53.1} & \textbf{53.1} \\
            DomainNet-126  & 33.9 & 33.1 & 32.8 & \textbf{32.7} & \textbf{32.7} \\
            \hline
            \end{tabular}}
        &
        \scalebox{0.93}{
            \tabcolsep4pt
            \begin{tabular}{l|ccccccc}\hline
              & 100\% & 50\% & 25\% & 10\% & 5\% & 1\% & 0\% \\
            \hline
            CIFAR10C      & \textbf{13.9} & \textbf{13.9} & 14.0 & 14.1 & 14.3 & 14.3 & 14.5 \\
            CIFAR100C     & \textbf{27.6} & 27.8 & 27.8 & 28.2 & 28.3 & 28.9 & 29.0 \\
            ImageNet-C    & 57.9 & \textbf{57.8} & 57.9 & \textbf{57.8} & 58.0 & 58.4 & 59.8 \\
            ImageNet-R    & 55.5 & 55.4 & \textbf{55.2} & 55.4 & 55.7 & 55.6 & 55.7 \\
            DomainNet-126 & \textbf{33.9} & 34.2 & 34.3 & 34.4 & 34.6 & 34.6 & 34.7 \\
            \hline
            \end{tabular}}\\
        \\
        (c) & (d) \\
                \scalebox{0.93}{
            \tabcolsep4pt
            \begin{tabular}{l|ccccc}\hline
             temperature $\tau$ & 0.01 & 0.07 & 0.1 & 0.2 & 1.0 \\
            \hline
            CIFAR10C      & \textbf{13.9} & \textbf{13.9} & \textbf{13.9} & \textbf{13.9} & 14.3 \\
            CIFAR100C     & \textbf{27.5} & \textbf{27.5} & 27.6 & 27.7 & 28.1 \\
            ImageNet-C    & \textbf{57.7} & 57.8 & 57.9 & \textbf{57.7} & 58.1 \\
            ImageNet-R    & 55.6 & \textbf{55.3} & 55.5 & 55.5 & 55.8 \\
            DomainNet-126 & 91.8 & 34.1 & \textbf{33.9} & 34.4 & 35.6 \\
            \hline
            \end{tabular}}
            &
                    \scalebox{0.93}{
            \tabcolsep4pt
            \begin{tabular}{l|ccccc}\hline
             momentum $\alpha$ & 0.99 & 0.995 & 0.999 & 0.9995 & 0.9999 \\
            \hline
            CIFAR10C      & \textbf{13.8} & \textbf{13.8} & 13.9 & 14.3 & 15.5 \\
            CIFAR100C     & 28.6 & 27.7 & \textbf{27.6} & 28.2 & 29.2 \\
            ImageNet-C    & 64.1 & 60.9 & \textbf{57.9} & 58.3 & 60.9 \\
            ImageNet-R    & 60.1 & 58.9 & \textbf{55.5} & 56.1 & 57.2 \\
            DomainNet-126 & 34.6 & 33.9 & \textbf{33.9} & 34.5 & 35.0 \\
            \hline
            \end{tabular}}
    \end{tabular}
    \label{tab:ablations_combined}
\end{table*}

\section{DomainNet-126}
\label{sec:appendix-domainnet}

While a variety of corruption benchmarks for test-time adaptation exist, benchmarks that investigate natural shifts are limited. A common dataset which contains natural shifts is ImageNet-R. However, it has the drawback that the included shifts are not separated. The continual DomainNet-126 benchmark closes this gap, consisting of four domains (real, clipart, painting, sketch) and 126 classes. It enables the investigation of continual TTA for natural shifts and further provides source models for all covered domains. Additionally, DomainNet-126 also includes shifts in label priors (imbalanced data), which corresponds to a more realistic setting than the uniform class distributions prevailing in existing benchmarks.

The exact four sequences used in the continual DomainNet-126 benchmark are shown in \cref{tab:domainnet_sequence_description}. While the left column indicates the name of the sequence and the domain on which the model was pre-trained, the order of the test domains is shown on the right.

Detailed results for the continual DomainNet-126 benchmark are shown in \cref{tab:domainnet126_detailed}.

\begin{table}[H] 
\renewcommand{\arraystretch}{1.2}
\centering
\caption{Details on the test sequences used for the continual DomainNet-126 benchmark.} \label{tab:domainnet_sequence_description}
\scalebox{0.93}{
\tabcolsep4pt
\begin{tabular}{l|lllll}\hline
 Source domain & \multicolumn{5}{c}{Test sequence} \\
\hline
real       & clipart & $\rightarrow$ & painting & $\rightarrow$ & sketch \\
clipart    & sketch & $\rightarrow$ & real & $\rightarrow$ & painting \\
painting   & real & $\rightarrow$ & sketch & $\rightarrow$ & clipart \\
sketch     & painting & $\rightarrow$ & clipart & $\rightarrow$ & real \\
\hline
\end{tabular}}
\end{table}

\begin{table*}[t] 
\renewcommand{\arraystretch}{1.2}
\centering
\caption{Classification error rate~(\%) for DomainNet-126 in the online continual test-time adaptation setting, where the test domains are sequentially displayed from left to right. We report the performance of our method averaged over 5 runs.} 
\label{tab:domainnet126_detailed}
\scalebox{0.93}{
\tabcolsep4pt
\begin{tabular}{l|c|c|ccc|ccc|ccc|ccc|c}\hline
\multicolumn{3}{l|}{Source domain} & \multicolumn{3}{c|}{real} & \multicolumn{3}{c|}{clipart} & \multicolumn{3}{c|}{painting} & \multicolumn{3}{c|}{sketch} \\
\hline
Method  & \rotatebox[origin=c]{90}{ Source-free } & \rotatebox[origin=c]{90}{ Updates } & clipart & painting & sketch & sketch & real & painting & real & sketch & clipart & painting & clipart & real & Mean \\
\hline
 Source & \ding{51} & - & 44.9 & 37.4 & 53.6 & 52.5 & 40.1 & 55.3 & 24.7 & 53.8 & 46.7 & 49.2 & 44.8 & 40.4 & 45.3 \\
 BN--1        & \ding{51} & - & 46.0 & 37.2 & 52.1 & 50.7 & 35.1 & 49.7 & 25.1 & 47.7 & 45.8 & 40.9 & 40.6 & 32.0 & 41.9 \\
 TENT cont.   & \ding{51} & 1 & 44.6 & 35.0 & 47.5 & 49.3 & 34.9 & 48.3 & 24.2 & 44.5 & 43.0 & 40.1 & 39.5 & 33.0 & 40.3 \\
 CoTTA        & \ding{51} & 1 & 45.3 & 35.8 & 49.2 & 50.1 & 33.4 & 45.6 & 23.4 & 43.9 & 41.8 & 40.0 & 39.2 & 29.6 & 39.8 \\
 AdaContrast  & \ding{51} & 1 & 39.3 & 32.7 & 41.4 &  45.5 & \textbf{27.5} & 39.7 & 20.5 & 39.2 & 37.2 & 35.9 & 34.7 & \textbf{25.1} & 34.9 \\
 GTTA-MIX     & \ding{55} & 4 & 40.4 & 32.7 & 42.8 & 46.4 & 34.2 & 46.7 & 23.3 & 40.2 & 37.3 & 36.9 & 36.0 & 29.8 & 37.7 \\
 RMT (ours)   & \ding{51} & 1 & 37.7 & 31.7 & 41.5 & 43.8 & 29.8 & 40.1 & 20.9 & 38.4 & 35.8 & 35.3 & 33.4 & 27.5 & 34.7 \\
 RMT (ours)   & \ding{55} & 1 & 37.9 & 31.4 & 41.1 & 43.5 & 29.1 & 38.6 & 21.1 & 37.1 & 33.8 & 35.1 & 31.7 & 26.4 & 33.9 \\
 RMT (ours)   & \ding{55} & 4 & \textbf{36.9} & \textbf{30.0} & \textbf{38.4} & \textbf{41.7} & 29.6 & \textbf{38.0} & \textbf{19.9} & \textbf{36.9} & \textbf{32.9} & \textbf{33.2} & \textbf{29.7} & 26.8 & \textbf{32.8} \\
\hline
\end{tabular}}
\end{table*}

\begin{figure*}[t]
  \centering
      \begin{tabular}{cc}
        \includegraphics[width=3.3in]{figures/cross-entropy.pdf} &
        \includegraphics[width=3.3in]{figures/symmetric-cross-entropy.pdf} \\
        \includegraphics[width=3.3in]{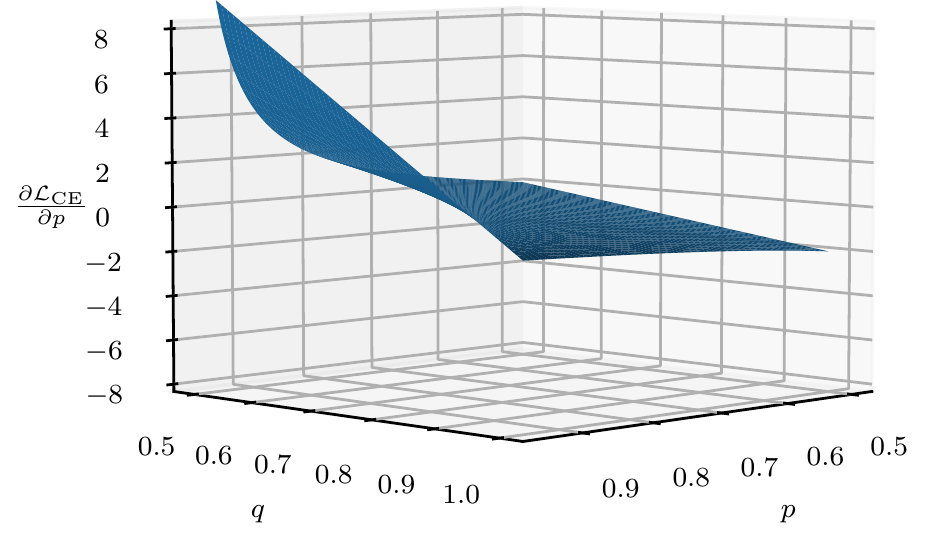} &
        \includegraphics[width=3.3in]{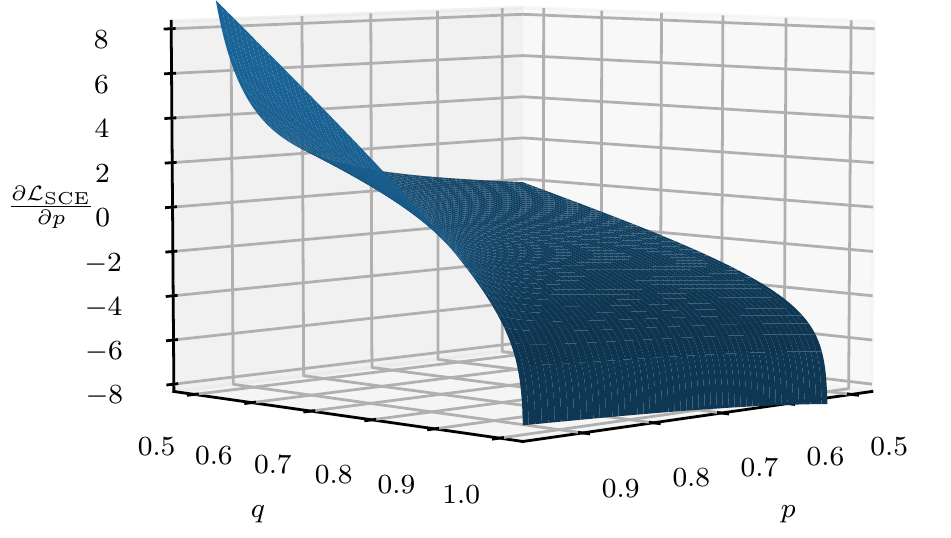}
      \end{tabular}
  \caption{Loss (top) and gradient (bottom) surface illustrated for the binary case of the cross-entropy loss $\mathcal{L}_\mathrm{CE}$ and the symmetric cross-entropy $\mathcal{L}_\mathrm{SCE}$ in dependence of the confidences $p$ and $q$ of the student and teacher, respectively.}
  \label{fig:grad_surface}
\end{figure*}

\end{document}